\documentclass{article}
\PassOptionsToPackage{table}{xcolor}

\usepackage[preprint]{neurips_2025}

\usepackage[utf8]{inputenc} 
\usepackage[T1]{fontenc}    
\usepackage{hyperref}       
\usepackage{url}            
\usepackage{booktabs}       
\usepackage{amsfonts}       
\usepackage{nicefrac}       
\usepackage{microtype}      
\usepackage{xcolor}         
\usepackage{amsmath}
\usepackage{enumitem}
\usepackage{graphicx}
\usepackage{amssymb}
\usepackage{xspace}
\usepackage{multirow}
\usepackage{wrapfig}
\usepackage{subcaption}
\usepackage[ruled,vlined]{algorithm2e}
\usepackage{titletoc}

\usepackage[table]{xcolor}   
\definecolor{LightBlue}{HTML}{E6EEF9}   
\definecolor{LightOrange}{HTML}{FFF2E6} 
\definecolor{LightGrey}{HTML}{F5F5F5}   

\definecolor{ForestGreen}{HTML}{228B22}
\definecolor{BrickRed}{HTML}{CB4154}

\newcommand{\diff}[1]{%
  \ifdim #1 pt > 0pt
    {\color{ForestGreen}\scriptsize(+\num{#1})}%
  \else
    {\color{BrickRed}\scriptsize(\num{#1})}%
  \fi
}

\usepackage{siunitx}

\newcommand{\ours}{FoReaL-Decoding\xspace}

\title{What makes Reasoning Models Different? Follow the Reasoning Leader for Efficient Decoding}

\author{ 
  \textbf{Ming Li}\textsuperscript{1,2}, 
  \textbf{Zhengyuan Yang}\textsuperscript{2}, 
  \textbf{Xiyao Wang}\textsuperscript{1}, 
  \textbf{Dianqi Li}, 
  \textbf{Linjie Li}\textsuperscript{2},
  \textbf{Kevin Lin}\textsuperscript{2}, \\
  \textbf{Tianyi Zhou}\textsuperscript{1} , 
  \textbf{Lijuan Wang}\textsuperscript{2}
  \\
  \textsuperscript{1}University of Maryland~~~~
  \textsuperscript{2}Microsoft \\
 minglii@umd.edu
}

\begin{document}

\maketitle

\begin{abstract}

Large reasoning models (LRMs) achieve strong reasoning performance by emitting long chains of thought. Yet, these verbose traces slow down inference and often drift into unnecessary detail, known as the overthinking phenomenon. 
To better understand LRMs' behavior, we systematically analyze the token-level misalignment between reasoning and non-reasoning models. 
While it is expected that their primary difference lies in the stylistic ``thinking cues'', LRMs uniquely exhibit two pivotal, previously under-explored phenomena: a \textbf{\textit{Global Misalignment Rebound}}, where their divergence from non-reasoning models persists or even grows as response length increases, and more critically, a \textbf{\textit{Local Misalignment Diminish}}, where the misalignment concentrates at the ``thinking cues'' each sentence starts with but rapidly declines in the remaining of the sentence. 
Motivated by the \textit{Local Misalignment Diminish}, 
we propose \textbf{\textit{FoReaL-Decoding}}, a collaborative fast-slow thinking decoding method for cost-quality trade-off. 
In \ours, a Leading model leads the first few tokens for each sentence, and then a weaker draft model completes the following tokens to the end of each sentence. \ours adopts a stochastic gate to smoothly interpolate between the small and the large model. On four popular math-reasoning benchmarks (AIME24, GPQA-Diamond, MATH500, AMC23), \ours reduces theoretical FLOPs by $30$ – $50\%$ and trims CoT length by up to $40\%$, while preserving $86$ – $100\%$ of model performance. These results establish \ours as a simple, plug-and-play route to controllable cost-quality trade-offs in reasoning-centric tasks.\looseness-1


\end{abstract}

\section{Introduction}

Reasoning has become a pivotal capability of large language models (LLMs), driving rapid progress in mathematical problem solving, code generation, and commonsense question answering \citep{huang2023reasoninglargelanguagemodels, li2024happened, ahn-etal-2024-large,wang2024mementos, wang2025multimodalchainofthoughtreasoningcomprehensive}. Contemporary Large Reasoning Models (LRMs) such as OpenAI’s GPT-o1 \citep{openai2024o1} and the open-source DeepSeek-R1 \citep{deepseekai2025deepseekr1incentivizingreasoningcapability} demonstrate this trend by producing explicit long chains of thought (CoT) \citep{wei2023chainofthoughtpromptingelicitsreasoning} that markedly improve performance on challenging tasks in mathematics \citep{xiong2025selfrewardingcorrectionmathematicalreasoning, xia2025evaluatingmathematicalreasoningaccuracy}, programming \citep{liu2024codemindframeworkchallengelarge}, and other complex domains. These deeper, longer, and more precise reasoning trajectories are cultivated by reinforcement-learning-based optimization \citep{deepseekai2025deepseekr1incentivizingreasoningcapability} or supervised fine-tuning on expert demonstrations \citep{ye2025limoreasoning, muennighoff2025s1simpletesttimescaling, li2025instruction}, representing advanced ``slow-thinking'' patterns \citep{kahneman2011thinking, li2024happened, li202512surveyreasoning}.
Although these slow-thinking LRMs showcase impressive reasoning skills, communities are increasingly concerned about the efficiency and fidelity of their often-lengthy chains of thought, a phenomenon known as ``overthinking'' \citep{chen2025think23overthinkingo1like, fan2025missing}, where excessive computational resources are allocated for simple problems with minimal benefit.

To alleviate overthinking and improve efficiency, a series of methods has been proposed \citep{yu2024distilling, team2025kimi, aggarwal2025l1controllinglongreasoning, xia2025tokenskip, luo2025o1prunerlengthharmonizingfinetuningo1like, hao2024training, shen2025codi, shen2025efficient, zhang2025lightthinker, han2024token, xu2025chain, renze2024benefits, sun2024fast, wan2024dynamic, wu2025more}.
Most of these, however, require further post-training or manipulate the LRM's distribution itself, adding complexity or computational overhead.
Motivated by Speculative Decoding \citep{leviathan2023fast} and the distinctions between fast and slow thinking, we ask: \textit{Is it possible to design a collaborative, training-free decoding method that mixes fast and slow thinking models to effectively trade-off quality and efficiency?}

To answer this and develop such a method, we first seek to pinpoint what truly differentiates strong reasoning models from standard instruction-following LLMs at the token level. 
For instruction-following models, LIMA \citep{zhou2023lima} proposes the ``superficial alignment'' hypothesis, in which it shows that most of the knowledge has been learned in the pretraining and only a small amount of data is needed for alignment. 
Although a line of work tries to use various methods for data selection on either instruction-following \citep{chen2023alpagasus, cherry, liu2023makes, Li2024SuperfilteringWD} or reasoning \citep{muennighoff2025s1simpletesttimescaling, ye2025limoreasoning} capabilities, \citep{lin2023unlocking} verifies this hypothesis from token-level analysis between the base model and the aligned model. 

Leveraging the diagnostic framework of \citep{lin2023unlocking}, our systematic analysis of misalignment across various model types (large reasoning, small reasoning, instruction-following, and pretrained base model) reveals critical insights.
We observe a ``superficial alignment'' phenomenon similar to \citep{lin2023unlocking}, where misaligned tokens are predominantly stylistic (e.g., ``\textit{Hmmm}'', ``\textit{Wait}'', ``\textit{Let me check}'') rather than content-specific, often related to explicit thinking patterns.
More strikingly, while previous work showed that misalignment between instruction-following and base models diminishes with longer context, 
we find this does not hold for reasoning models.
Instead, we identify a \textbf{\textit{Global Misalignment Rebound}}, where overall misalignment between reasoning and non-reasoning models can slightly grow with response length, suggesting that increasing the length cannot reduce the misalignment.
This indicates that the reasoning abilities are \textit{not} as superficial as instruction-following.
Crucially, despite this global trend, we uncover a corresponding \textbf{\textit{Local Misalignment Diminish}} phenomenon: most token misalignments occur at the \textit{beginning of each sentence}, then rapidly decrease until the next sentence starts.
These findings reveal a novel \textit{periodical, sentence-level misalignment diminishing pattern} unique to LRMs, driven by thinking-pattern indicators concentrated at sentence openings, shedding light on a better understanding of token-level divergences of these two types of models. 


Based on this core insight that the reasoning pattern of LRMs is often front-loaded in each sentence, we hypothesize that strategic, limited intervention by a strong LRM can guide a weaker model, balancing reasoning quality with efficiency.
To this end, we propose \textbf{\textit{\underline{Fo}llow the \underline{Rea}soning \underline{L}eader (\ours)}}, an efficient collaborative decoding method. 
In \ours, a strong Leading model generates the initial few tokens of each sentence (capturing the potentially misaligned ``thinking cues''), after which a weaker Draft model completes the sentence.
To further mitigate potential overthinking from the Leading model (e.g., endlessly generating ``Wait''), we introduce a stochastic binary gate that controls whether the Leading model intervenes on a given sentence.
These two control knobs, lead token count and lead probability, allow \ours to smoothly interpolate between the Draft and Leading models, offering strong controllability over the cost-quality spectrum.

\paragraph{Contributions.}
In summary, our primary contributions can be illustrated as follows:
\begin{itemize}[leftmargin=1em]

\item We conduct a systematic token-level analysis comparing LRMs with non-reasoning models, identifying two pivotal, under-explored phenomena:
(1) \textbf{\textit{Global Misalignment Rebound}}, where the token distribution of LRMs diverges from that of non-reasoning models and their gap even increases with longer responses; 
(2) \textbf{\textit{Local Misalignment Diminish}}, where LRMs only make noticeable difference 
on generating stylistic ``thinking-patterns'' at the very beginning of each sentence. But such divergence from non-reasoning models rapidly drops on subsequent tokens within the sentence. This periodical \textit{sentence-level misalignment diminishing} pattern has not been explored previously.
These two discoveries significantly advance the understanding of LRMs.



\item Leveraging these analytical insights (particularly the \textit{Local Misalignment Diminish}), we propose \textbf{\textit{\ours{}}}, a training-free, collaborative algorithm that mixes the strength of a ``slow-thinking'' LRM (as Leading model) with the efficiency of a ``fast-thinking'', weaker model (as draft model). \ours{} is designed to be plug-and-play, offering strong controllability to balance the cost and quality under diverse budgets of tokens. 

\item Experimental results on several reasoning-heavy math tasks (AIME24, GPQA-Diamond, MATH500, AMC23) demonstrate that \ours{} reduces FLOPS by $30$-$55$\% and CoT length by up to $40$\%, while preserving $86$-$100$\% of the leading model's performance, effectively mitigating ``overthinking''.

\end{itemize}

\section{Token Distributions of Reasoning vs. Non-Reasoning Models}

\begin{figure*}[t]
\centering 
\includegraphics[width=1.0\textwidth]{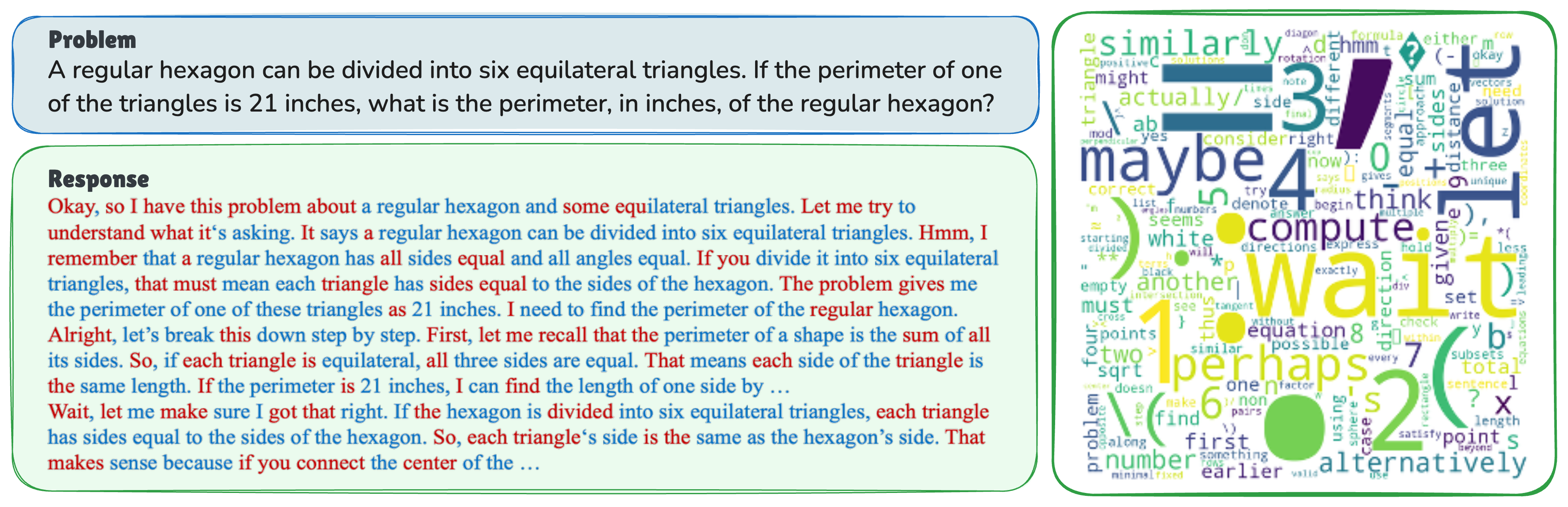} 
\caption{
\textbf{Left}: An example comparing the token distribution alignment between \textit{DeepSeek-R1-Distill-Qwen-32B} and \textit{Qwen2.5-1.5B-Instruct}, qualitatively showing that the misaligned tokens (colored in \textcolor{red}{red}) are related to thinking patterns, and probably appear at the start of sentences.  
\textbf{Right}: The WordCloud of the misaligned tokens calculated on a mix of math datasets, quantitatively showing the high-frequency misaligned tokens like ``\textit{wait}'', ``\textit{perhaps}'', ``\textit{maybe}'', ``\textit{let}'', and ``\textit{alternatively}''. 
} 
\vspace{-4mm}
\label{fig:analysis_example} 
\end{figure*}


Large-scale reasoning models (LRMs) often outperform smaller instruction-tuned models on complex reasoning-heavy tasks, yet how their generation behavior differs from instruction models within the same model family remains unclear. 
\citep{lin2023unlocking} proposes an analytical method through the lens of token-distribution shifts and finds that alignments between instruction-following and base pretrained models are often superficial. 
This phenomenon is supported by nearly identical decoded tokens in the majority of token positions under the same input contexts, with distribution shifts occurring mainly with stylistic tokens like discourse markers. 
However, the critical question remains: ``\textit{Does this superficial alignment finding on instruction-following LLMs still hold for today's capable LRMs?}'' 
Thus, our work systematically investigates token misalignment across various model combinations involving LRMs. 


\paragraph{Experimental Setup \& Metric.}
In this analysis, we utilize \textit{DeepSeek-R1-Distill-Qwen-32B} as the targeting LRM, which we notate as the Leading model $P_L(\cdot)$. The corresponding small models, within the same family, that are used for comparison are noted as the Draft models $P_D(\cdot)$. 
The Draft models can be (i) the pretrained base model (\textit{Qwen2.5-1.5B}), (ii) the instruction-following model (\textit{Qwen2.5-1.5B-Instruct}), or (iii) the small reasoning model (\textit{DeepSeek-R1-Distill-Qwen-1.5B}) in our analysis and method. 
For a user query $q$, the output response generated greedily from the Leading model can be notated as $y=\{y_{1}, ..., y_{T} \}$, where $T$ represents the length of the response. 
This response serves as the target for calculating the token distribution for the Draft model. 
At each position $t$, the context for predicting this token can be presented as $c_t=<q; y_{<t}>$, where $<;>$ represents the concatenation operation.

In the analysis, the aligned positions are defined as those token steps where the Draft model, when conditioned on the Leading model’s history, would greedily generate exactly the same token as the Leading model, which means that \textit{the two models have the same most probable behavior under the same context, indicating the alignment}. 

Suppose $\mathcal{V}$ is the vocabulary for next-token prediction, then the aligned token indices are:
\begin{equation}
\mathcal{A}
\;=\;
\Bigl\{
\,t \in \{1,\dots,T\}\;:\;
\underset{y\in\mathcal{V}}{\arg\max}\,
P_{D}\!\bigl(y \mid c_t\bigr)
=\,
\underset{y\in\mathcal{V}}{\arg\max}\,
P_{L}\!\bigl(y \mid c_t\bigr)
\Bigr\},
\end{equation}
which collects exactly those positions where the Draft model’s top-1 prediction matches the Leading model’s emitted token under the shared causal context $c_t$. Thus, the aligned and misaligned tokens can be defined: 
\begin{equation}
\mathbf{y}_{\mathcal{A}} \;=\; \{\, y_t \mid t \in \mathcal{A} \}
\qquad
\mathbf{y}_{\mathcal{A^{\complement}}} \;=\; \{\, y_t \mid t \notin \mathcal{A} \}
\end{equation}


\paragraph{Qualitative Analysis on Misaligned Tokens.}
Figure \ref{fig:analysis_example} (left) shows a qualitative example (truncated) from MATH500, comparing the token distribution alignment between \textit{DeepSeek-R1-Distill-Qwen-32B} as the Leading model and \textit{Qwen2.5-1.5B-Instruct} as the Draft model. The shown response $y$ is generated by the Leading model, the aligned tokens $\mathbf{y}_{\mathcal{A}}$ are colored in blue, and misaligned tokens $\mathbf{y}_{\mathcal{A^{\complement}}}$ are colored in red.
Through the example, it can be intuitively perceived that the misaligned tokens are mostly stylistic tokens related to thinking patterns, and the beginning of each sentence has a larger probability of being misaligned. 
To further quantitatively investigate what exactly these misaligned tokens are, we extract all the misaligned tokens from the mix of AIME24, AMC23, GPQA, and MATH datasets, count their frequencies, and generate the corresponding WordCloud shown in Figure \ref{fig:analysis_example} (right). 
From the WordCloud, it is observed that most of the high-frequency misaligned tokens are related to thinking patterns of LRMs, like ``\textit{wait}'', ``\textit{perhaps}'', ``\textit{maybe}'', ``\textit{let}'', and ``\textit{alternatively}'', which shows a similar but different superficial phenomenon than previous instruction-following LLMs: While misalignment in both types of models is primarily stylistic rather than content-based, those in LRMs are distinctively characterized by tokens reflecting their overt reasoning or self-correction patterns.
Thus, our qualitative exploration reveals that LRM misalignment is characterized by stylistic ``thinking cues'' concentrated at sentence beginnings, prompting a more detailed quantitative analysis of their underlying distribution patterns.



\begin{wrapfigure}[35]{r}{0.5\textwidth}
\vspace{-10pt} 
\centering
\includegraphics[width=1.0\linewidth]{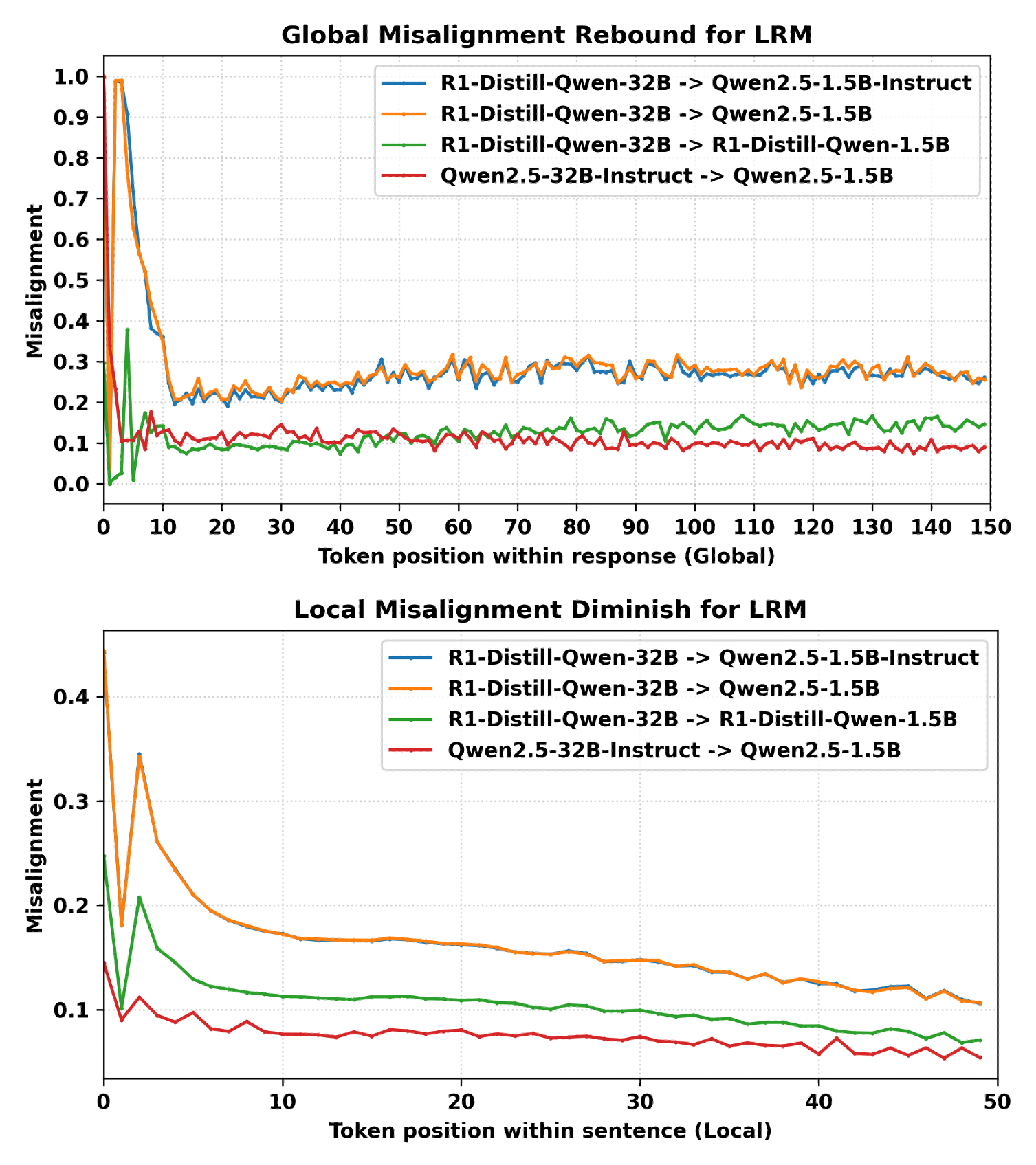}
\captionsetup{width=\linewidth}
\caption{\textbf{Top}: Response-level misalignment changes with response length. \textbf{Bottom}: sentence-level misalignment changes with response length. The y-axis represents the average misalignment rate at each token position, the x-axis represents the token position within the whole response or sentence. We reveal the novel Global Misalignment Rebound and Local Misalignment Diminish phenomenon that only occurs on current LRMs, shown as the \textcolor{blue}{blue}, \textcolor{orange}{orange}, and \textcolor{green}{green} lines of the upper figure. This phenomenon does not hold for the previous alignment between the instruction-following and base models, shown in the \textcolor{red}{red} line.\looseness-1} 
\label{fig:analysis_figure}
\vspace{-20pt} 
\end{wrapfigure}

\vspace{-4mm}
\paragraph{Global Misalignment Rebound.}
Existing analysis on token distribution shifts between instruct and base models has identified that such shifts will gradually diminish over time during the decoding process due to the more comprehensive context given, as shown in Figure~\ref{fig:analysis_figure} (upper, red line).
In the figure, the y-axis represents the average misalignment rate at each token position, while the x-axis represents the token position within the whole response (upper panel) or sentence (lower panel).
As shown, the red line, representing misalignment between the instruct model and base model, decreases and remains at a low rate.
This implies that providing longer context can gradually compensate for the misalignment between instruct and base models.

However, this response-level misalignment diminishing phenomenon does not strictly hold for LRMs.
As illustrated in Figure~\ref{fig:analysis_figure} (upper), lines corresponding to LRM as the Leading model exhibit different behaviors. When the Draft models are instruct (blue line) or base (orange line) models, the misalignment rates initially decrease dramatically to around $0.2$, then rebound and persist around $0.3$.
In contrast, the green line, representing misalignment between large and small reasoning models (which belong to the same family and are trained on similar data), shows consistently low misalignment from the beginning, indicating a distinct trend.
We term the observed persistent or rebounding divergence between LRMs and non-reasoning models the \textbf{\textit{Global Misalignment Rebound}} phenomenon.
This phenomenon, characteristic of LRM comparisons with non-reasoning models, is mainly caused by LRMs continuously generating thinking patterns at the beginning of sentences, partly to prevent premature conclusion of the generation process.
This finding demonstrates that merely extending context length is insufficient to resolve the misalignment between reasoning and non-reasoning models, indicating that reasoning capability is \textit{not} as superficial as instruction-following. 


\vspace{-3mm}
\paragraph{Local Misalignment Diminish.}
It is uncommon that a longer context does not benefit the alignment. 
Thus, to further understand this behavior, we conduct the sentence-level analysis by calculating the token misalignment rate at each sentence-level position. 
In the response, sentences can be separated by periods, question marks, exclamation marks, and the newline symbol. 
Specifically, for any position $x$, we first collect every sentence that is at least $x$ tokens long. Mark the $x$-th token in each of those sentences as $1$ if it is misaligned and $0$ if it is aligned. The average of these $0$-$1$ indicators across all selected sentences is the misalignment rate for position $x$.

As shown in Figure \ref{fig:analysis_example} (lower), for the red line, there is no obvious misalignment decrease that can be observed. 
It means that between the instruct and the base model, the misalignment occurs relatively evenly across the whole sentence. 
On the contrary, for LRM-involved model combinations, the blue, orange, and green lines, the misalignment rates drop dramatically at the first several tokens, e.g., from $0.4$ to $0.15$, and then keep diminishing, indicating a totally different behavior. 
Thus, we term this phenomenon the \textit{\textbf{Local Misalignment Diminish}} phenomenon for reasoning models.  
These findings reveal a novel periodical, sentence-level misalignment diminish pattern unique to LRMs, driven by thinking-pattern indicators concentrated at sentence openings, shedding light on a better understanding of token-level divergences of these two types of models. 


\vspace{-2mm}
\paragraph{Findings.}
From this section, several key findings can be concluded:
\begin{itemize}[leftmargin=1em]

    \item LRM misalignment with non-reasoning models, while largely superficial and characterized by stylistic ``thinking cues'', uniquely exhibits a \textbf{\textit{Global Misalignment Rebound}}. Unlike instruct models that increasingly align with more context, token divergence at the response level can persist or even grow, underscoring deeper, ingrained differences in their generative behavior.

    \item LRMs distinctively display a \textbf{\textit{Local Misalignment Diminish}}. This manifests as a novel, periodical sentence-level pattern where high misalignment, driven by ``thinking cues'' concentrated at sentence beginnings, rapidly decreases as the sentence progresses. This predictable intra-sentence dynamic is a crucial insight for developing LRM-guided decoding and understanding LRM patterns.

\end{itemize}

\vspace{-2mm}
\section{\ours}
\vspace{-2mm}

Motivated by the above token divergence analysis, we propose a collaborative fast-slow thinking decoding method for cost-quality Trade-off, \textbf{\textit{\underline{Fo}llow the \underline{Rea}soning \underline{L}eader (\ours)}}, a plug-and-play training-free method that mixes the strength of a slow but highly capable large reasoning model with the speed of a small model. 
The central idea is to let the strong, large (\emph{Leading}) model lead at the beginning of sentences, and allow the weaker, small (\emph{Draft}) model to complete the rest of the tokens. 
This decoding algorithm is of strong controllability, which can smoothly transfer into the Leading model only or downgrade to the Draft model only, by controlling the probability and the number of tokens to lead. 



\paragraph{Preliminaries.}

The two control knobs that govern the trade-off between cost and quality:
\begin{enumerate}[leftmargin=1.5em,itemsep=2pt]
    \item \textbf{Required lead count} \(n\in\mathbb{N}\): the minimum number of tokens the Leading model generates before yielding control to the Draft model.
    \item \textbf{Lead probability} \(p\in[0,1]\): probability that a sentence is led by the Leading model. 
\end{enumerate}
When \(p=0\), the decoding system degenerates to pure Draft model decoding; when \(p=1\) and \(n\) exceeds the sentence length, it transfers to Leading model decoding. Intermediate settings form a continuity of compute–accuracy trade-offs.

In addition, let \(t\in\mathbb{N}\) represent the global token index in the response, and \(s\in\mathbb{N}\) represent the sentence index. \(g_s\sim\mathrm{Bernoulli}(p)\) 
represents the sentence-level gate to decide what model to start the sentence $s$: the sentence will be led by the Leading model if $g_s=1$. 
\(\tau_s\) represents the global position of the first token in $s$. 
\(s(t)=\max\{s:\tau_s\le t\}\) is the function that maps the token $t$ to the sentence index that $t$ belongs to. 
\(\lambda_t=t-\tau_{s(t)}+1\) is the local position of token $t$ within its sentence. 


\paragraph{Intra-Sentence Lead}

Within a sentence $s$, the generation of each token at position $t$ is governed by the token-level policy, 

\begin{equation}
\pi_t \;=\;
\begin{cases}
L &
\!g_{s(t)}=1\;\land\;
\bigl[\lambda_t\le n \;\lor\; t<H^{\mathrm{hit}}_{s(t)}\bigr],
\\[4.5pt]
D & \text{otherwise}.
\end{cases}
\label{eq:policy}
\end{equation}

$g_{s(t)}=1$ represents this sentence $s(t)$ should be led by the Leading model, decided by the gate. 
$L$ and $D$ represent the Leading model and Draft model, respectively. 
$\lambda_t\le n$ represents the index of this token within this sentence that is smaller than the required lead count $n$, thus should be generated by the Leading model. 
\(H^{\mathrm{hit}}_s\) is the first token index within $s$ where the top-$1$ token generated by the Draft model matches that of the Leading model for $k$ consecutive steps:

\begin{equation}
H^{\mathrm{hit}}_s=\min\{t:\,s(t)=s,\,\lambda_t>n,\,h_t=k\},
\end{equation}
where $h_t$ represents the number of consecutive hits within the max sliding window of $k$:  

\begin{equation}
h_t=\sum_{i=0}^{k-1}\delta_{t-i},\quad
\delta_t=\mathbf{1}\!\bigl\{\underset{y\in\mathcal{V}}{\arg\max} P_D(\cdot|c_t)=\underset{y\in\mathcal{V}}{\arg\max} P_L(\cdot|c_t)\bigr\}
\end{equation}

Put it simply, for each sentence, if the Bernoulli gate decides to let $P_L$ lead the sentence with the probability $p$, $P_L$ will generate the first $n$ tokens. Then, $P_D$ begins the generation process as well, with the purpose of measuring the alignment between the two models. 
When the top-$1$ predictions of these two models aligned with each other for $k$ times, the generation process is yielded to $P_D$, otherwise, $P_L$ generates the whole sentence. On the contrary, if the gate decides not to let $P_L$ lead, then the whole sentence will be completely generated by $P_D$.

\paragraph{Sentence-level likelihood.}
For sentence $s$ with token span $Y_s=(y_{\tau_s},\dots,y_{\tau_{s+1}-1})$ and length $L_s$, the conditional likelihood under \ours is:

\begin{equation}
P_{\text{CoL}}\!\bigl(Y_s \mid g_s\bigr)
  \;=\;
  \prod_{i=0}^{L_s-1}
  P_{\pi_{\tau_s+i}}\!\bigl(
        y_{\tau_s+i}\mid c_{\tau_s+i}
     \bigr),
\label{eq:sent-lik-pi}
\end{equation}

Whenever \(\pi_t=L\), the factor draws its probability from the distribution \(P_L\) of the Leader model; otherwise from the Draft model of distribution \(P_D\).

\paragraph{Inter-Sentence Transfer}

Upon encountering a sentence boundary at the token \(t\), i.e., the sentence is complete, we execute the inter-sentence update by resetting the hit counter and resampling the gate for the next sentence.  
\begin{equation}
s \leftarrow s+1,\quad
g_s \sim \mathrm{Bernoulli}(p),\quad
h_t\leftarrow 0
\end{equation}







\paragraph{Comparisons with related methods.}
In speculative decoding \citep{leviathan2023fast}, the final text provably matches what the large model alone would have produced. 
However, our \textbf{\textit{\ours}} focuses on the reasoning-heavy scenarios where the responses generated by the LRM itself are not desirable due to the overthinking. 
Thus, our method serves as a deliberate mixture of two distributions, aiming at reducing the overthinking problem of LRMs by inserting the distribution from weaker models, and at the same time increasing the efficiency.  
A recent work, RSD \citep{liao2025reward}, also aims at reducing computation cost by utilizing speculative decoding. However, it introduces an additional process reward model as the judge, while our method focuses on utilizing the collaborative models themselves only, thus, it is largely different from our settings. 
Another concurrent work, Speculative Thinking \citep{yang2025speculativethinkingenhancingsmallmodel}, also shares similar motivation as ours, in which a ``small-writes, large-fixes'' mechanism is utilized, which differs from our ``large-leads, small-follows''. Moreover, \textbf{\textit{\ours}} provides a smooth transition from the small to the large model, representing wider trade-off scopes. 



\vspace{-2mm}
\section{Experiments}
\vspace{-2mm}

\subsection{Implementation Details}

\paragraph{Models, Datasets, and Setup.}

To assess the effects of \ours, extensive experiments are conducted for different model combinations in the Qwen2.5 family, including reasoning models like \textit{R1-Distill-Qwen-32B}~\citep{deepseekai2025deepseekr1incentivizingreasoningcapability},
\textit{R1-Distill-Qwen-1.5B}~\citep{deepseekai2025deepseekr1incentivizingreasoningcapability}, non-reasoning instruct models like \textit{Qwen2.5-7B-Instruct}~\citep{qwen2.5}, \textit{Qwen2.5-1.5B-Instruct}~\citep{qwen2.5}, and base models like \textit{Qwen2.5-1.5B}~\citep{qwen2.5}. To cover a wide scope of potential trade-offs, we utilize the reasoning models as the Leading models, while any of the above types as the Draft models. 
Moreover, our extensive experiments on the recently released \textit{Qwen3}~\citep{qwen3} series further verify the generalizability of our method. 
We evaluate our method on relatively hard, reasoning-heavy math datasets, including AIME2024~\citep{aime2024}, GPQA-Diamond~\citep{rein2024gpqa}, AMC23~\citep{amc23}, and MATH500~\citep{lightman2023lets}.
All experiments were conducted on NVIDIA A100 GPUs (80G), utilizing the Huggingface Transformers package. During the generation, we follow the recommended generation configuration from R1-Distill models as \texttt{temperature=$0.6$}, \texttt{top\_p=$0.95$}, \texttt{top\_k=$40$} for all the experiments. During the generation, we always let the Leading model generate the first paragraph, and we fix the required hits for generation transfer as $k=5$ for all the experiments.

\begin{table*}[t]
  \centering
  \caption{Comparisons of Accuracy and Efficiency (TFLOPs) of \ours on commonly used reasoning-heavy math problem tasks. To further show the wide trade-off scopes of our method, we provide some different configurations as the control. The results of Speculative Thinking are the reported results. The accuracies are better with higher ($\uparrow$) values, while the TFLOPs are better with lower ($\downarrow$) values. The accuracies on each line are compared with the Draft model, and the TFLOPs are compared with the Leading models: better values are colored in {\color{ForestGreen} green}, otherwise {\color{BrickRed} red}. }
  \label{tab:main}
  \resizebox{\textwidth}{!}{%
  \begin{tabular}{ lc *{4}{cc} }
  \toprule
  \multicolumn{2}{c}{\textbf{Model}} &
  \multicolumn{2}{c}{\textbf{AIME24}} &
  \multicolumn{2}{c}{\textbf{GPQA-D}} &
  \multicolumn{2}{c}{\textbf{MATH500}} &
  \multicolumn{2}{c}{\textbf{AMC23}} \\
  \cmidrule(lr){1-2}\cmidrule(lr){3-4}\cmidrule(lr){5-6}\cmidrule(lr){7-8}\cmidrule(lr){9-10}
  \textbf{Method} & \textbf{Config} & \textbf{ACC (\%) $\uparrow$} & \textbf{TFLOPs $\downarrow$} & \textbf{ACC (\%) $\uparrow$} & \textbf{TFLOPs $\downarrow$} & \textbf{ACC (\%) $\uparrow$} & \textbf{TFLOPs $\downarrow$} & \textbf{ACC (\%) $\uparrow$} & \textbf{TFLOPs $\downarrow$} \\
  \midrule
  \multicolumn{10}{l}{\textbf{DeepSeek-R1-Distill-Qwen-32B + DeepSeek-R1-Distill-Qwen-1.5B}}\\
  \midrule
  \multicolumn{2}{l}{DeepSeek-R1-Distill-Qwen-32B} & 66.7 & 15.72 & 59.6 & 8.09 & 93.6 & 4.13 & 95.0 & 7.54 \\
  \multicolumn{2}{l}{DeepSeek-R1-Distill-Qwen-1.5B} & 23.3 & 2.86 & 22.2 & 1.13 & 81.4 & 1.14 & 65.0 & 2.51 \\
  \multicolumn{2}{l}{Speculative Thinking} & 32.2 & 5.75 & 41.9 & 2.62 & 89.4 & 1.51 & 80.0 & 3.31 \\
  \rowcolor{LightBlue}
  \ours & $n$=15,$p$=0.4 & 33.3~{\color{ForestGreen}\scriptsize(+10.0)} & \textbf{5.60~{\color{ForestGreen}\scriptsize(-10.12)}} & 43.3~{\color{ForestGreen}\scriptsize(+21.1)} & \textbf{2.47~{\color{ForestGreen}\scriptsize(-5.62)}} & 90.2~{\color{ForestGreen}\scriptsize(+8.8)} & \textbf{1.43~{\color{ForestGreen}\scriptsize(-2.88)}} & 80.0~{\color{ForestGreen}\scriptsize(+15.0)} & \textbf{2.91~{\color{ForestGreen}\scriptsize(-4.63)}} \\
  \rowcolor{LightBlue}
  \ours & $n$=15,$p$=0.6 & 50.0~{\color{ForestGreen}\scriptsize(+26.7)} & 6.77~{\color{ForestGreen}\scriptsize(-8.95)} & 48.2~{\color{ForestGreen}\scriptsize(+26.0)} & 4.50~{\color{ForestGreen}\scriptsize(-3.59)} & 91.4~{\color{ForestGreen}\scriptsize(+10.0)} & 2.40~{\color{ForestGreen}\scriptsize(-1.26)} & 80.0~{\color{ForestGreen}\scriptsize(+15.0)} & 3.99~{\color{ForestGreen}\scriptsize(-3.55)} \\
  \rowcolor{LightBlue}
  \ours & $n$=15,$p$=0.8 & 50.0~{\color{ForestGreen}\scriptsize(+26.7)} & 8.47~{\color{ForestGreen}\scriptsize(-7.25)} & 54.6~{\color{ForestGreen}\scriptsize(+32.4)} & 4.69~{\color{ForestGreen}\scriptsize(-3.40)} & 93.4~{\color{ForestGreen}\scriptsize(+12.0)} & 2.70~{\color{ForestGreen}\scriptsize(-1.43)} & 90.0~{\color{ForestGreen}\scriptsize(+25.0)} & 5.37~{\color{ForestGreen}\scriptsize(-2.17)} \\
  \rowcolor{LightBlue}
  \ours & $n$=15,$p$=1.0 & \textbf{66.7~{\color{ForestGreen}\scriptsize(+43.4)}} & 9.16~{\color{ForestGreen}\scriptsize(-6.56)} & 56.6~{\color{ForestGreen}\scriptsize(+34.4)} & 6.21~{\color{ForestGreen}\scriptsize(-1.88)} & 93.2~{\color{ForestGreen}\scriptsize(+11.8)} & 3.14~{\color{ForestGreen}\scriptsize(-0.99)} & 92.5~{\color{ForestGreen}\scriptsize(+27.5)} & 5.28~{\color{ForestGreen}\scriptsize(-2.26)} \\
  \rowcolor{LightOrange}
  \ours & $n$=25,$p$=0.8 & 53.3~{\color{ForestGreen}\scriptsize(+30.0)} & 10.95~{\color{ForestGreen}\scriptsize(-4.77)} & 57.1~{\color{ForestGreen}\scriptsize(+34.9)} & 5.65~{\color{ForestGreen}\scriptsize(-2.44)} & 92.6~{\color{ForestGreen}\scriptsize(+11.2)} & 3.13~{\color{ForestGreen}\scriptsize(-1.00)} & 92.5~{\color{ForestGreen}\scriptsize(+27.5)} & 4.99~{\color{ForestGreen}\scriptsize(-2.55)} \\
  \rowcolor{LightOrange}
  \ours & $n$=25,$p$=1.0 & \textbf{66.7~{\color{ForestGreen}\scriptsize(+43.4)}} & 10.54~{\color{ForestGreen}\scriptsize(-5.18)} & \textbf{57.6~{\color{ForestGreen}\scriptsize(+35.4)}} & 6.68~{\color{ForestGreen}\scriptsize(-1.41)} & \textbf{94.5~{\color{ForestGreen}\scriptsize(+13.1)}} & 3.50~{\color{ForestGreen}\scriptsize(-0.63)} & \textbf{95.0~{\color{ForestGreen}\scriptsize(+30.0)}} & 5.66~{\color{ForestGreen}\scriptsize(-1.88)} \\
  \midrule
  \multicolumn{10}{l}{\textbf{DeepSeek-R1-Distill-Qwen-32B + Qwen2.5-1.5B-Instruct}}\\
  \midrule
  \multicolumn{2}{l}{DeepSeek-R1-Distill-Qwen-32B} & 66.7 & 15.72 & 59.6 & 8.09 & 93.6 & 4.13 & 95.0 & 7.54 \\
  \multicolumn{2}{l}{Qwen2.5-1.5B-Instruct} & 0.0 & 0.12 & 23.7 & 0.12 & 49.2 & 0.09 & 15.0 & 0.10 \\
  \rowcolor{LightBlue}
  \ours & $n$=15,$p$=0.8 & 20.0~{\color{ForestGreen}\scriptsize(+20.0)} & \textbf{9.05~{\color{ForestGreen}\scriptsize(-6.67)}} & 38.4~{\color{ForestGreen}\scriptsize(+14.7)} & 5.63~{\color{ForestGreen}\scriptsize(-2.46)} & 76.2~{\color{ForestGreen}\scriptsize(+27.0)} & 2.85~{\color{ForestGreen}\scriptsize(-1.28)} & 65.0~{\color{ForestGreen}\scriptsize(+50.0)} & 5.22~{\color{ForestGreen}\scriptsize(-2.32)} \\
  \rowcolor{LightBlue}
  \ours & $n$=15,$p$=1.0 & 20.0~{\color{ForestGreen}\scriptsize(+20.0)} & 11.19~{\color{ForestGreen}\scriptsize(-4.53)} & 47.5~{\color{ForestGreen}\scriptsize(+23.8)} & 5.86~{\color{ForestGreen}\scriptsize(-2.23)} & 85.9~{\color{ForestGreen}\scriptsize(+36.7)} & 3.28~{\color{ForestGreen}\scriptsize(-0.85)} & 85.0~{\color{ForestGreen}\scriptsize(-70.0)} & 6.15~{\color{ForestGreen}\scriptsize(-1.39)} \\
  \rowcolor{LightOrange}
  \ours & $n$=25,$p$=0.8 & 36.7~{\color{ForestGreen}\scriptsize(+36.7)} & 9.58~{\color{ForestGreen}\scriptsize(-6.14)} & 45.0~{\color{ForestGreen}\scriptsize(+21.3)} & \textbf{4.37~{\color{ForestGreen}\scriptsize(-3.72)}} & 82.0~{\color{ForestGreen}\scriptsize(+32.8)} & \textbf{2.52~{\color{ForestGreen}\scriptsize(-1.61)}} & 72.5~{\color{ForestGreen}\scriptsize(+57.5)} & \textbf{4.65~{\color{ForestGreen}\scriptsize(-2.89)}} \\
  \rowcolor{LightOrange}
  \ours & $n$=25,$p$=1.0 & \textbf{40.0~{\color{ForestGreen}\scriptsize(+40.0)}} & 11.00~{\color{ForestGreen}\scriptsize(-4.72)} & \textbf{57.1~{\color{ForestGreen}\scriptsize(+33.4)}} & 6.27~{\color{ForestGreen}\scriptsize(-1.82)} & \textbf{90.8~{\color{ForestGreen}\scriptsize(+2.8)}} & 3.36~{\color{ForestGreen}\scriptsize(-0.77)} & \textbf{92.5~{\color{ForestGreen}\scriptsize(-77.5)}} & 6.88~{\color{ForestGreen}\scriptsize(-0.66)} \\
  \midrule
  \multicolumn{10}{l}{\textbf{DeepSeek-R1-Distill-Qwen-1.5B + Qwen2.5-7B-Instruct}}\\
  \midrule
  \multicolumn{2}{l}{DeepSeek-R1-Distill-Qwen-1.5B} & 23.3 & 2.86 & 22.2 & 1.13 & 81.4 & 1.14 & 65.0 & 2.51 \\
  \multicolumn{2}{l}{Qwen2.5-7B-Instruct} & 6.7 & 0.95 & 38.4 & 0.89 & 76.0 & 0.61 & 52.5 & 0.75 \\
  \multicolumn{2}{l}{Speculative Thinking} & 6.7 & 4.93 & 31.8 & 6.73 & 74.8 & 2.04 & 55.0 & 4.97 \\
  \rowcolor{LightBlue}
  \ours & $n$=15,$p$=0.8 & 16.7~{\color{ForestGreen}\scriptsize(+10.0)} & 2.05~{\color{ForestGreen}\scriptsize(-0.81)} & \textbf{34.3~{\color{BrickRed}\scriptsize(-4.1)}} & 1.07~{\color{ForestGreen}\scriptsize(-0.06)} & 76.4~{\color{ForestGreen}\scriptsize(+0.4)} & 0.57~{\color{ForestGreen}\scriptsize(-0.57)} & 57.5~{\color{ForestGreen}\scriptsize(+5.0)} & \textbf{1.08~{\color{ForestGreen}\scriptsize(-1.43)}} \\
  \rowcolor{LightBlue}
  \ours & $n$=15,$p$=1.0 & 16.7~{\color{ForestGreen}\scriptsize(+10.0)} & 6.47~{\color{BrickRed}\scriptsize(+3.61)} & 29.8~{\color{BrickRed}\scriptsize(-8.6)} & 3.08~{\color{BrickRed}\scriptsize(+1.95)} & \textbf{79.6~{\color{ForestGreen}\scriptsize(+3.6)}} & 1.42~{\color{BrickRed}\scriptsize(+0.28)} & 52.5~{\color{ForestGreen}\scriptsize(+0.0)} & 3.35~{\color{BrickRed}\scriptsize(+0.84)} \\
  \rowcolor{LightOrange}
  \ours & $n$=25,$p$=0.8 & 20.0~{\color{ForestGreen}\scriptsize(+13.3)} & \textbf{1.57~{\color{ForestGreen}\scriptsize(-1.29)}} & 33.3~{\color{BrickRed}\scriptsize(-5.1)} & \textbf{0.80~{\color{ForestGreen}\scriptsize(-0.33)}} & 78.6~{\color{ForestGreen}\scriptsize(+2.6)} & \textbf{0.55~{\color{ForestGreen}\scriptsize(-0.59)}} & \textbf{65.0~{\color{ForestGreen}\scriptsize(+12.5)}} & 1.76~{\color{ForestGreen}\scriptsize(-0.75)} \\
  \rowcolor{LightOrange}
  \ours & $n$=25,$p$=1.0 & \textbf{23.3~{\color{ForestGreen}\scriptsize(+16.6)}} & 3.18~{\color{BrickRed}\scriptsize(+0.32)} & 29.3~{\color{BrickRed}\scriptsize(-9.1)} & 2.53~{\color{BrickRed}\scriptsize(+1.40)} & 79.2~{\color{ForestGreen}\scriptsize(+3.2)} & 1.04~{\color{ForestGreen}\scriptsize(-0.10)} & \textbf{65.0~{\color{ForestGreen}\scriptsize(+12.5)}} & 1.66~{\color{ForestGreen}\scriptsize(-0.85)} \\
  \bottomrule
  \end{tabular}}%
  \vspace{-4mm}
\end{table*}


\subsection{Main Results}

Table \ref{tab:main} presents the comparisons between accuracy and efficiency (TFLOPs) of \ours on commonly used reasoning-heavy math problem tasks. We provide some different configurations as controls to show the wide trade-off scopes of our method. We also present the reported results of the concurrent work, Speculative Thinking \citep{yang2025speculativethinkingenhancingsmallmodel}, for better comparison. The accuracies on each line are compared with the Draft model, and the TFLOPs are compared with the Leading models: better values are colored in green, otherwise red. We utilize the theoretically estimated TFLOPs as the efficiency measurement since it takes the generation length into account, different from the estimated speed. In the main comparison, we focus on 3 collaborative settings. Across four benchmarks, \ours cuts inference cost by $30$ – $55\%$ relative to Leader-only decoding while retaining $86$ – $100\%$ of its accuracy. The detailed statistics, including response length and leading ratios on AIME24, can be found in Table \ref{tab:table_detail_1} for better understanding.

\textit{R1-Distill-Qwen-32B for Leading, R1-Distill-Qwen-1.5B for Draft.}
This collaborative setting yields the highest accuracies for all of the math reasoning datasets. In this setting, the larger $32$B reasoning model takes charge of the leading of the sentences, while the smaller $1.5$B reasoning model needs to complete the remaining sentence. In this setting, both models have the reasoning capabilities, but \ours implicitly separates the generation of each sentence into two phases and yields the less informative Draft phase to the smaller model for better efficiency. 
As shown in the table, all our results obtain better performances compared with the Draft model and efficiencies compared with the Leading model, and also exceed Speculative Thinking, indicating the capability of our methods. 
Moreover, on all of the tasks except GPQA-D, \ours reaches similar or even slightly higher performances than the $32$B Leading model with fewer TFLOPs, ($-6.56$ on AIME24, $-0.63$ on MATH500, and $-1.88$ on AMC). 

\textit{R1-Distill-Qwen-32B for Leading, Qwen2.5-1.5B-Instruct for Draft.} 
This setting represents a direct mixture of a large reasoning model and a small non-reasoning model. As shown in the table, the $1.5$B instruct model performs badly on the given difficult math problems. The use of a stronger reasoning model for leading largely improves the accuracy, although with more computation required. The response lengths are largely shorter than \textit{R1-Distill-Qwen-1.5B}, representing an alleviation of overthinking. 
Compared with using another small reasoning model for Draft, utilizing the instruction model leads to suboptimal performance. To understand this phenomenon, further experiments are conducted where the base pretrained model \textit{Qwen2.5-1.5B } is utilized as the Draft model. As shown in Table \ref{tab:table_detail_1}, the accuracies, response lengths, and TFLOPs are almost identical compared with using base and instruct models, which means the previous instruction-aligned process does not benefit the current reasoning settings.

\begin{wraptable}[24]{r}{0.52\textwidth}
  \vspace{-12pt}
  \caption{The detailed results of different collaborative settings on AIME. Length represents the averaged response length, Ratio represents the average ratio of tokens decoded by using the Leading model, for each task. Additional configuration that uses base model for Draft is included. }
  \centering
  \resizebox{0.5\textwidth}{!}{%
  \begin{tabular}{ lc c c c c }
    \toprule
    \multicolumn{2}{c}{\textbf{Model}} &
    \multicolumn{4}{c}{\textbf{AIME24}}\\
    \cmidrule(lr){1-2}\cmidrule(lr){3-6}
    \textbf{Method} & \textbf{Config} & \textbf{ACC (\%)} & \textbf{Length} & \textbf{Ratio} & \textbf{TFLOPs} \\
    \midrule
    \multicolumn{6}{l}{\textbf{DeepSeek-R1-Distill-Qwen-32B + DeepSeek-R1-Distill-Qwen-1.5B}}\\
    \midrule
    FoReaL-Decoding & $n{=}15,\,p{=}0.4$ & 33.3 & 11\,876 & 0.272 & 5.60 \\
    FoReaL-Decoding & $n{=}15,\,p{=}0.6$ & 50.0 & 10\,934 & 0.401 & 6.77 \\
    FoReaL-Decoding & $n{=}15,\,p{=}0.8$ & 50.0 & 11\,532 & 0.527 & 8.47 \\
    FoReaL-Decoding & $n{=}15,\,p{=}1.0$ & 66.7 & 10\,617 & 0.666 & 9.16 \\
    FoReaL-Decoding & $n{=}25,\,p{=}0.8$ & 53.3 & 12\,081 & 0.676 & 10.95 \\
    FoReaL-Decoding & $n{=}25,\,p{=}1.0$ & 66.7 & 11\,116 & 0.683 & 10.54 \\
    \midrule
    \multicolumn{6}{l}{\textbf{DeepSeek-R1-Distill-Qwen-32B + Qwen2.5-1.5B-Instruct}}\\
    \midrule
    FoReaL-Decoding & $n{=}15,\,p{=}0.8$ & 20.0 & 12\,584 & 0.571 & 9.05 \\
    FoReaL-Decoding & $n{=}15,\,p{=}1.0$ & 20.0 & 14\,188 & 0.588 & 11.19 \\
    FoReaL-Decoding & $n{=}25,\,p{=}0.8$ & 36.7 & 11\,575 & 0.710 & 9.58 \\
    FoReaL-Decoding & $n{=}25,\,p{=}1.0$ & 40.0 & 11\,239 & 0.813 & 11.00 \\
    \midrule
    \multicolumn{6}{l}{\textbf{DeepSeek-R1-Distill-Qwen-32B + Qwen2.5-1.5B (Base)}}\\
    \midrule
    FoReaL-Decoding & $n{=}15,\,p{=}0.8$ & 23.3 & 12\,224 & 0.547 & 9.56 \\
    FoReaL-Decoding & $n{=}15,\,p{=}1.0$ & 20.0 & 12\,107 & 0.664 & 10.39 \\
    \midrule
    \multicolumn{6}{l}{\textbf{DeepSeek-R1-Distill-Qwen-1.5B + Qwen2.5-7B-Instruct}}\\
    \midrule
    FoReaL-Decoding & $n{=}15,\,p{=}0.8$ & 16.7 & 4\,120 & 0.545 & 2.05 \\
    FoReaL-Decoding & $n{=}15,\,p{=}1.0$ & 16.7 & 14\,132 & 0.651 & 6.47 \\
    FoReaL-Decoding & $n{=}25,\,p{=}0.8$ & 20.0 & 4\,474 & 0.693 & 1.57 \\
    FoReaL-Decoding & $n{=}25,\,p{=}1.0$ & 23.3 & 11\,436 & 0.841 & 3.18 \\
    \bottomrule
  \end{tabular}}%
  \label{tab:table_detail_1}
  \vspace{-3pt}
\end{wraptable}

\textit{R1-Distill-Qwen-1.5B for Leading, Qwen2.5-7B-Instruct for Draft.}
Different from the above settings, in which a strong but large reasoning model is used as the Leading model, this setting considers a different and most efficient scenario, utilizing a small reasoning model for leading and a slightly larger instruct model for Draft. In this setting, the efficiencies are reduced to an extremely low level, even faster than directly utilizing the small reasoning models. As shown in Table \ref{tab:table_detail_1}, \ours largely reduces the length required for the problem, thus largely reducing the computation required. On AIME24 and AMC23, our method reaches the same accuracy as the Leading model with similar or less computation. On GPQA, our method reaches an intermediate accuracy, since the abnormal situation where a non-reasoning model has better performance than the reasoning model. 

\vspace{-4mm}
\paragraph{Estimation of TFLOPs.}
Empirical latency depends on vendor-specific kernel fusion and memory layouts, so a timing measured on one backend may not transfer to another. Counting floating-point operations (FLOPs) provides a hardware-agnostic yardstick that isolates algorithmic differences. The performance figures we report are presented in TeraFLOPs (TFLOPs), where one TFLOP equals $10^{12}$ FLOPs.
Typically, the generation process proceeds in two modes, prefill and decode. Prefill processes the full prompt of length $s$ once without any KV cache, and decode autoregressively emits output tokens while re-using cached keys and values. When GPU memory is sufficient, profiling shows that producing multiple tokens during the prefix phase costs almost the same as decoding a single token. Therefore, we upper-bound the prefix cost by the single-token decode cost. For the TFLOPs values cited in our results, we calculate the precise total FLOPs using the detailed formulas presented in Appendix A.1. This calculation methodology is based on~\citep{chen2024towards,han2024inference}, and the resulting total FLOPs are then converted to TFLOPs for reporting.

\begin{figure*}[t]
\centering 
\includegraphics[width=1.0\textwidth]{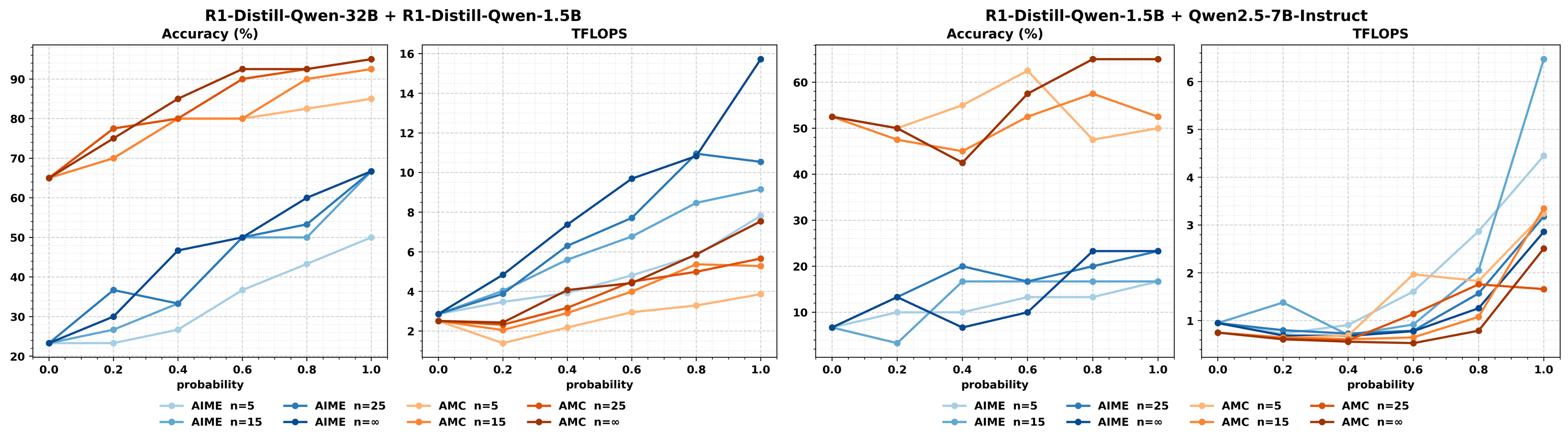} 
\caption{
Effects of lead count and lead probability on AIME24 and AMC23 datasets, based on 2 collaborative configurations. \ours provides a smooth cost-quality trade-off, making the transition from the weak Draft model to the strong Leading model smooth and controllable. 
} 
\label{fig:trade_off_1} 
\vspace{-4mm}
\end{figure*}

\vspace{-2mm}
\subsection{Effects of Lead Count and Lead Probability}
\vspace{-2mm}

Figure \ref{fig:trade_off_1} sweeps the two hyperparameters that govern the controllability of \ours, lead count $n$ and lead probability $p$ on AIME24 and AMC23 datasets, based on 2 collaborative configurations, \textit{DeepSeek-R1-Distill-Qwen-32B + DeepSeek-R1-Distill-Qwen-1.5B} and \textit{DeepSeek-R1-Distill-Qwen-1.5B + Qwen2.5-7B-Instruct}, representing the high-performance and high-efficiency settings, respectively. For each model combination, we run experiments on $n \in \{5,15,25,+\infty\}, p \in \{0,0.2,0.4,0.6,0.8,1.0\}$. When $p=0$, \ours utilizes the Draft model only, and utilizes the Leading model only when $p=1$ and $n=+\infty$. 
According to the figure, \ours provides a smooth cost-quality trade-off, making the transition from the weak Draft model to the strong Leading model smooth and controllable. 
For any fixed~$n$, increasing the probability~$p$ of the Leader intervention shifts the operating point up and to the right: accuracy
rises while TFLOPs grow almost linearly. The resulting curve is smooth, allowing practitioners to trade latency for quality by adjusting
$(n,p)$. The jump from $n=5$ to $n=15$ yields large accuracy gains at a modest cost increase. Further enlarging the Leader count ($n\ge25$) adds little accuracy yet inflates compute up a lot, indicating that sentence-level guidance already captures most of the benefit of slow reasoning. A sweet spot is around $(\,n,p\,)=\,(15,0.6)$.

\vspace{-2mm}
\subsection{Trade-Off Curves}
\vspace{-2mm}

\begin{wrapfigure}[16]{r}{0.5\textwidth}
\vspace{-10pt} 
\centering
\includegraphics[width=1.0\linewidth]{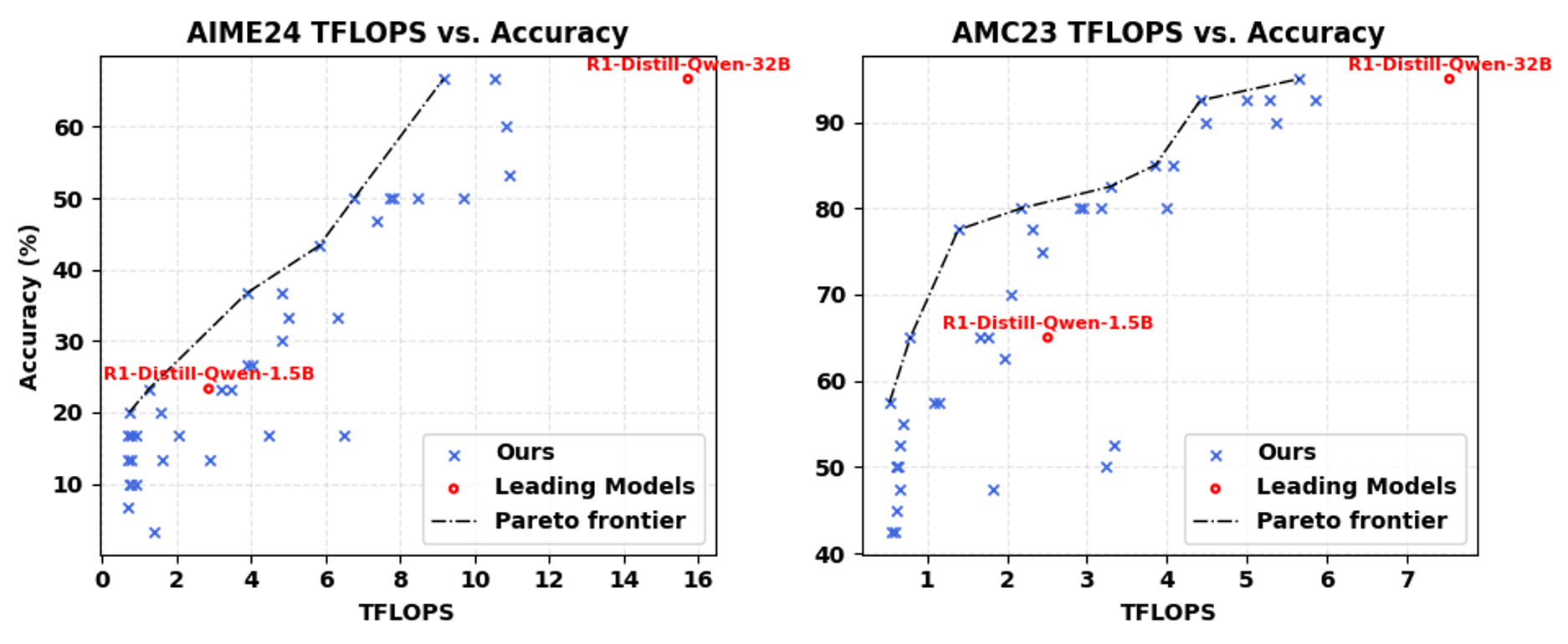}
\captionsetup{width=\linewidth}
\caption{The trade-off curves between accuracy and TFLOPs. Blue markers correspond to \ours variants, red circles denote the corresponding LRMs, and the dashed line is the empirically computed Pareto frontier. On both benchmarks, every LRM point is Pareto dominated. } 
\label{fig:trade_off_2}
\vspace{-20pt} 
\end{wrapfigure}

Figure \ref{fig:trade_off_2} plots the trade-off curves between accuracy and TFLOPs for every $(n,p)$ configuration tested on AIME24 (left) and AMC23 (right), according to the experiment scopes from the above section. Blue markers correspond to \ours variants, red circles denote the corresponding LRMs, and the dashed line is the empirically computed Pareto frontier. 
On both benchmarks, every LRM point is Pareto dominated: an alternative \ours setting always achieves higher accuracy at lower cost.
Moreover, we find that the frontier rises sharply between $0.5$ and $2$ TFLOPs, as each additional TFLOP yields $10$–$15$ percentage points of accuracy. However, beyond ${\approx}5$ TFLOPs, the curve flattens; extra compute buys only marginal improvements up to the ceiling.  

\vspace{-2mm}
\subsection{Results on Qwen3 Families}
\vspace{-2mm}


\begin{wraptable}[15]{r}{0.5\textwidth}
  \vspace{-15pt}
  \caption{The detailed results of Qwen3 series models on AIME. \ours shows promising performance in this additional family. }
  \centering
  \resizebox{0.5\textwidth}{!}{%
  \begin{tabular}{ lc c c c c }
    \toprule
    \multicolumn{2}{c}{\textbf{Model}} &
    \multicolumn{4}{c}{\textbf{AIME24}}\\
    \cmidrule(lr){1-2}\cmidrule(lr){3-6}
    \textbf{Method} & \textbf{Config} & \textbf{ACC (\%)} & \textbf{Length} & \textbf{Ratio} & \textbf{TFLOPs} \\
    \midrule
    \multicolumn{6}{l}{\textbf{Base Models}}\\
    \midrule
    Qwen3-32B      & --                & 76.6 & 13\,275 & --    & 15.75 \\
    Qwen3-1.7B     & --                & 40.0 & 14\,990 & --    & 2.81 \\
    Qwen3-0.6B     & --                & 13.3 & 15\,839 & --    & 1.14 \\
    \midrule
    \multicolumn{6}{l}{\textbf{Qwen3-32B + Qwen3-1.7B}}\\
    \midrule
    FoReaL-Decoding & $n{=}15,\,p{=}0.4$ & 60.0 & 14\,840 & 0.272 & 7.20 \\
    FoReaL-Decoding & $n{=}15,\,p{=}0.6$ & 73.3 & 14\,110 & 0.412 & 8.83 \\
    FoReaL-Decoding & $n{=}15,\,p{=}0.8$ & 73.3 & 15\,081 & 0.536 & 11.43 \\
    \midrule
    \multicolumn{6}{l}{\textbf{Qwen3-32B + Qwen3-0.6B}}\\
    \midrule
    FoReaL-Decoding & $n{=}15,\,p{=}0.4$ & 36.7 & 17\,782 & 0.281 & 7.44 \\
    FoReaL-Decoding & $n{=}15,\,p{=}0.6$ & 63.0 & 14\,279 & 0.410 & 8.18 \\
    FoReaL-Decoding & $n{=}15,\,p{=}0.8$ & 60.0 & 15\,478 & 0.560 & 11.01 \\
    \bottomrule
  \end{tabular}}%
  \label{tab:table_other}
  \vspace{-6pt}
\end{wraptable}

To further verify the effectiveness and generalizability of \ours, additional experiments are conducted on the Qwen3 series of models, including \textit{Qwen3-32B}, \textit{Qwen3-1.7B}, and \textit{Qwen3-0.6B}, due to the various sizes of models provided in the family. Specifically, we utilize the reasoning modes for these models and follow exactly the same generation configuration for our main experiments. 
The detailed experimental results are shown in Table \ref{tab:table_other}. As shown in the table, \ours shows promising performances on both \textit{Qwen3-32B + Qwen3-1.7B} and \textit{Qwen3-32B + Qwen3-0.6B} configurations. For the former configuration, \ours reaches a similar accuracy ($73.3\%$ to $76.6\%$) with approximately half of the TFLOPs ($8.83$ to $15.75$).


\section{Conclusion, Limitation, and Future Direction}


Our systematic token-level analysis comparing Large Reasoning Models (LRMs) with non-reasoning models has uncovered two pivotal, previously under-explored divergence phenomena. First, we identified a \textbf{\textit{Global Misalignment Rebound}}, where LRM token divergence from non-reasoning models can unexpectedly persist or even increase over entire responses, underscoring deep-seated generative differences not easily bridged by extended context. Second, and critically for our method, we characterized the \textbf{\textit{Local Misalignment Diminish}}: a novel, periodical sentence-level pattern wherein LRM-specific stylistic ``thinking cues'' cause high token divergence at the very beginning of sentences, after which this misalignment rapidly decreases within the sentence.

Leveraging the predictable nature of the \textit{Local Misalignment Diminish}, we proposed \textbf{\textit{FoReaL-Decoding (\ours{})}}, a training-free, plug-and-play collaborative decoding algorithm. \ours{} strategically allows a strong LRM to lead the crucial initial tokens of sentences (capturing these divergent ``thinking cues''), while a lightweight Draft model efficiently completes the subsequent, more aligned portions. A stochastic gate further modulates the LRM's intervention to mitigate overthinking and control the cost-quality trade-off. Our experiments demonstrate that \ours{} achieves significant efficiency gains, reducing theoretical TFLOPs.

Current limitations include the manual definition of lead count ($n$) and lead probability ($p$).  Future work could explore adaptive mechanisms for determining $n$ and $p$ dynamically, or investigate possibilities for bidirectional feedback within the collaborative decoding framework to enhance reasoning fidelity further.


\clearpage
\bibliography{custom}
\bibliographystyle{plain}


\clearpage
\appendix
\startcontents[appendix]
\setcounter{tocdepth}{1}
\printcontents[appendix]{ }{0}{\section*{Table of Contents for Appendix}}

\clearpage

\section{Pseudo Code}

The pseudo code of our \ours is provided below, all the variables are kept the same as in the main context. 

\begin{algorithm}[htbp]
\caption{FoReaL-Decoding}
\label{alg:foreal}
\KwIn{Leading model ${P}_L$, Draft model ${P}_D$, lead count $n$, lead probability $p$, hit threshold $k$, input prompt $q$, max new tokens $\text{MAX\_LEN}$}
\KwOut{Generated tokens $\mathbf{y}$}

$\mathbf{y} \gets []$, $h \gets 0$, $\lambda \gets 0$\;
$c \gets q$ \tcp*{Initial context}
$g \gets 1$ \tcp*{Initialize gate}

\While{$\text{len}(\mathbf{y}) < \text{MAX\_LEN}$}{
    \If{$\text{is\_sentence\_boundary}(\mathbf{y}[-1])$}{
        $g \sim \text{Bernoulli}(p)$ \tcp*{Sample gate for new sentence}
        $h \gets 0$ \tcp*{Reset hit counter}
        $\lambda \gets 0$ \tcp*{Reset position in sentence}
    }
    
    $\lambda \gets \lambda + 1$ \tcp*{Increment position in sentence}
    
    \tcp{Generate next token}
    \eIf{$g = 1$ \textbf{and} $(\lambda \leq n$ \textbf{or} $h < k)$}{
        $t \gets \text{sample}({P}_L(\cdot | c))$ \tcp*{Use Leading model}
    }{
        $t \gets \text{sample}({P}_D(\cdot | c))$ \tcp*{Use Draft model}
    }
    
    \tcp{Check alignment when approaching transition point}
    \If{$g = 1$ \textbf{and} $\lambda > n - k$}{ 
        \eIf{$\text{top-1}({P}_D(\cdot | c)) = \text{top-1}({P}_L(\cdot | c))$}{
            $h \gets h + 1$\;
        }{
            $h \gets 0$\;
        }
    }
    
    $\mathbf{y}.\text{append}(t)$\;
    $c \gets \text{concat}(c, t)$ \tcp*{Update context}
    
    \If{$t \in \text{EOS\_tokens}$}{
        \textbf{break}\;
    }
}

\Return{$\mathbf{y}$}\;
\end{algorithm}

\clearpage

\section{FLOPs Calculation}

The calculation of Floating Point Operations (FLOPs) for the prefill and decoding stages is based on the methodology from \citep{chen2024towards,han2024inference}. These calculations assume a batch size of 1.

The variables involved are defined as:
\begin{itemize}
    \item $s$: Represents the sequence length.
        \begin{itemize}
            \item For the prefill stage ($\text{FLOPs}_{\text{prefill}}(s)$), $s$ is the length of the input prompt, denoted as $p_l$.
            \item For the decode stage ($\text{FLOPs}_{\text{decode}}(s)$), $s$ is the current length of the context (prompt + tokens generated so far) that the model attends to via its Key-Value (KV) cache.
        \end{itemize}
    \item $h$: The hidden size of the model.
    \item $h'$: The intermediate size of the feed-forward network (FFN).
    \item $n$: The number of attention heads.
    \item $p_l$: The length of the initial problem prompt.
    \item $d_l$: The number of tokens to be generated in the solution.
\end{itemize}
It is noted that the hidden size $h$ relates to the number of attention heads $n$ and the size of each attention head $d$ by $h = n \cdot d$.

The FLOPs for the prefill stage, which processes the initial input prompt of length $s=p_l$, is given by Equation~\ref{eq:flops_prefill_appendix}:
\begin{equation}
    \text{FLOPs}_{\text{prefill}}(s) = 8sh^2 + 16sh + 4s^2h + 4s^2n + 6shh' + 2sh'
    \label{eq:flops_prefill_appendix}
\end{equation}

The FLOPs for the decode stage, which generates a single token when the current KV cache has a length of $s$, is given by Equation~\ref{eq:flops_decode_appendix}:
\begin{equation}
    \text{FLOPs}_{\text{decode}}(s) = 8h^2 + 16h + 4sh + 4sn + 6hh' + 2h'
    \label{eq:flops_decode_appendix}
\end{equation}

The total FLOPs to generate $d_l$ tokens from a prompt of length $p_l$ combines the prefill cost for the prompt and the sum of decode costs for each generated token, as shown in Equation~\ref{eq:flops_total_appendix}:
\begin{equation}
    \text{FLOPs}_{\text{total}} = \text{FLOPs}_{\text{prefill}}(p_l) + \sum_{i=0}^{d_l-1} \text{FLOPs}_{\text{decode}}(p_l + i)
    \label{eq:flops_total_appendix}
\end{equation}
In this formula, for the $i$-th token being generated (0-indexed), the argument to $\text{FLOPs}_{\text{decode}}$ is $p_l+i$, representing the sequence length in the KV cache at that generation step.

\clearpage

\section{Related Works}

\subsection{Large Reasoning Models}

Recent advances in large language models (LLMs) have spurred a surge of work aimed at strengthening their reasoning abilities~\citep{ahn-etal-2024-large, besta2025reasoninglanguagemodelsblueprint, chen2025reasoningerasurveylong}.
Core reasoning skills are already instilled during pre-training, where models absorb commonsense and mathematical patterns from vast text corpora~\citep{touvron2023llama2openfoundation,openai2024o1}.
Researchers have therefore concentrated on post-training techniques to further polish these skills. One prominent direction employs reinforcement learning to nudge models toward more effective chains of thought~\citep{shao2024deepseekmathpushinglimitsmathematical, xiong2025selfrewardingcorrectionmathematicalreasoning, cui2025processreinforcementimplicitrewards,wang2025sota}. Another line shows that carefully curated instruction-tuning data can likewise deliver sizable gains in reasoning accuracy~\citep{ye2025limoreasoning, muennighoff2025s1simpletesttimescaling,wang2024scaling}.

Despite the impressive benchmark scores of recent Reasoning Language Models, several studies have begun to probe the quality and efficiency of the reasoning they generate. \citep{xia2025evaluatingmathematicalreasoningaccuracy} conduct a broad assessment and reveal substantial redundancy in many model-produced solutions. Follow-up investigations~\citep{chen2025think23overthinkingo1like, cuadron2025dangeroverthinkingexaminingreasoningaction, qu2025surveyefficientreasoninglarge, liu2025efficientinferencelargereasoning, fan2025missing} underscore an “overthinking” phenomenon, whereby models craft unduly verbose derivations even for simple problems. Capitalizing on this trait, \citep{kumar2025overthinkslowdownattacksreasoning} demonstrate a slowdown attack: small input perturbations can trigger excessive reasoning, markedly degrading inference speed.

To alleviate overthinking and improve efficiency for reasoning models, a series of efficient reasoning methods has been proposed. For example, \citep{yu2024distilling, team2025kimi, aggarwal2025l1controllinglongreasoning, xia2025tokenskip, luo2025o1prunerlengthharmonizingfinetuningo1like} utilize model-based methods that either add further constraints on RL rewards or SFT on diverse lengths of CoTs, \citep{hao2024training, shen2025codi, shen2025efficient, zhang2025lightthinker} utilize latent-space reasoning methods that transfer the massive tokens into the embedding space, \citep{han2024token, xu2025chain, renze2024benefits} utilize the prompt-based methods, \citep{sun2024fast, wan2024dynamic, wu2025more} utilize the sampling methods. 
Most of these methods either require further post-training or manipulating the distribution of LRM itself. 

\subsection{Alignment and Token Pattern Analysis}

A key empirical foundation for LLM Alignment is LIMA \citep{zhou2023lima}, which demonstrated that just $1,000$ carefully curated instruction–response pairs are already enough for LLM alignment, crystallizing the ``superficial alignment'' hypothesis. 
While a line of work directly follows the hypotheses by introducing data selection or alignment methods \citep{chen2023alpagasus, cherry, li2023reflection, Li2024SelectiveRS, du2023mods, li2024ruler, bukharin2023data, liu2023makes, Li2024SuperfilteringWD, li2024mosaic, li2025instruction, Xu2024ASO}, there are also works that try to further investigate this phenomenon. 

\citep{lin2023unlocking} provides a comprehensive token-level evidence by comparing the top-k token distributions of base models and their chat-tuned counterparts. 
The authors show that almost all divergence concentrates on discourse markers, politeness phrases, and safety disclaimers, while core content tokens remain unchanged. 
\citep{chen2025extracting} dissects which prompt-level cues are sufficient (and which are not) for alignment, showing that reasoning gaps emerge precisely where superficial patterns end. The debate has sparked push-back as well: \citep{raghavendra2024revisiting} demonstrates systematic performance gains when the amount of post-training data scales up, arguing that some deeper representational changes do accrue beyond mere style. 
Researchers are also probing where superficial signals live: \citep{li2024superficial} argues that data curation, not extra optimization steps, is the primary lever: filtering for safety disclaimers yields larger alignment jumps than adding thousands of generic examples. 
Together, these works paint a nuanced picture: much of the alignment gap after pre-training is indeed ``superficial'', residing in a narrow band of stylistic tokens that can be manipulated through tiny prompts, judicious data selection. 
However, in this paper, we show that \textit{the reasoning capabilities might not be as superficial as previous findings.}

\subsection{Speculative Decoding and Collaborative Decoding}

Speculative decoding, inaugurated by \citep{leviathan2023fast}, uses a small ``draft'' model to propose several tokens that the large ``target'' model then verifies in one batch, yielding $2$–$3\times$ latency reductions with provably identical output distributions. Follow-up work, such as \citep{chen2023accelerating} extends the idea to 70 B-parameter models and confirms similar speed-ups, while \citep{cai2024medusa} replaces the external draft model with extra decoding heads to remove system complexity 
System-level schedulers like \citep{liu2024optimizing} dynamically adapt draft length to traffic conditions and push end-to-end gains beyond $3\times $ in production settings.

Collaborative decoding improves text quality by letting multiple models cooperate during generation. \citep{li2022contrastive} runs a weak ``amateur'' model alongside a strong ``expert'' and selects tokens that maximize their likelihood gap, sharply reducing repetition and incoherence without retraining. \citep{jin2024collaborative} introduces a critical-token strategy that switches to the pretrained base model whenever factual precision is needed, cutting hallucinations in instruction-tuned LLMs. At an even finer grain, \citep{shen2024learning} treats ``who should emit the next token'' as a latent variable, enabling on-the-fly delegation between a generalist LLM and domain specialists and outperforming any single model on cross-domain tasks.

For the recent models with strong reasoning capabilities, several recent works \citep{liao2025reward, yang2025speculativethinkingenhancingsmallmodel} based on speculative decoding have also been released, which we have introduced in the main method section. Our \ours provides a different method with a better trade-off scope. 

\clearpage

\section{Detailed Results}

Table \ref{tab:appendix_detailed_main} and Table \ref{tab:appendix_detailed_tradeoff} show the detailed results of different settings of our method.

\begin{table}[h]
  \caption{The detailed results of different collaborative settings on AIME24, GPQA-D, MATH500, and AMC23, including length and ratio.  }
  \centering
  \resizebox{\textwidth}{!}{%
  \begin{tabular}{ lc c c c c c c c c c c c c c c c c }
    \toprule
    \multicolumn{2}{c}{\textbf{Model}} &
    \multicolumn{4}{c}{\textbf{AIME24}} &
    \multicolumn{4}{c}{\textbf{GPQA-D}} &
    \multicolumn{4}{c}{\textbf{MATH500}} &
    \multicolumn{4}{c}{\textbf{AMC23}}\\
    \cmidrule(lr){1-2}\cmidrule(lr){3-6}\cmidrule(lr){7-10}\cmidrule(lr){11-14}\cmidrule(lr){15-18}
    \textbf{Method} & \textbf{Config} & \textbf{ACC (\%)} & \textbf{Length} & \textbf{Ratio} & \textbf{TFLOPs} & \textbf{ACC (\%)} & \textbf{Length} & \textbf{Ratio} & \textbf{TFLOPs} & \textbf{ACC (\%)} & \textbf{Length} & \textbf{Ratio} & \textbf{TFLOPs} & \textbf{ACC (\%)} & \textbf{Length} & \textbf{Ratio} & \textbf{TFLOPs} \\
    \midrule
    \multicolumn{18}{l}{\textbf{DeepSeek-R1-Distill-Qwen-32B + DeepSeek-R1-Distill-Qwen-1.5B}}\\
    \midrule
    \multicolumn{2}{l}{DeepSeek-R1-Distill-Qwen-32B} & 66.7 & 13035 & - & 15.72 & 59.6 & 6602 & - & 8.09 & 93.6 & 3542 & - & 4.13 & 95.0 & 6243 & - & 7.54 \\
    \multicolumn{2}{l}{DeepSeek-R1-Distill-Qwen-1.5B} & 23.3 & 18021 & - & 2.86 & 22.2 & 8696 & - & 1.13 & 81.4 & 6704 & - & 1.14 & 65.0 & 13311 & - & 2.51 \\
    \midrule
    \rowcolor{LightBlue} FoReaL-Decoding & $n{=}15,\,p{=}0.4$ & 33.3 & 11876 & 0.272 & 5.60 & 43.3 & 5841 & 0.294 & 2.47 & 90.2 & 3402 & 0.312 & 1.45 & 80.0 & 6043 & 0.304 & 2.91 \\
    \rowcolor{LightBlue}FoReaL-Decoding & $n{=}15,\,p{=}0.6$ & 50.0 & 10934 & 0.401 & 6.77 & 48.2 & 7007 & 0.431 & 4.50 & 91.4 & 3995 & 0.452 & 2.40 & 80.0 & 6460 & 0.429 & 3.99 \\
    \rowcolor{LightBlue}FoReaL-Decoding & $n{=}15,\,p{=}0.8$ & 50.0 & 11532 & 0.527 & 8.47 & 54.6 & 6110 & 0.570 & 4.69 & 93.4 & 3658 & 0.590 & 2.70 & 90.0 & 7037 & 0.571 & 5.37 \\
    \rowcolor{LightBlue}FoReaL-Decoding & $n{=}15,\,p{=}1.0$ & 66.7 & 10617 & 0.666 & 9.16 & 56.6 & 6796 & 0.692 & 6.21 & 93.2 & 3655 & 0.726 & 3.14 & 92.5 & 5942 & 0.708 & 5.28 \\
    \rowcolor{LightOrange}FoReaL-Decoding & $n{=}25,\,p{=}0.8$ & 53.3 & 12081 & 0.676 & 10.95 & 57.7 & 6223 & 0.702 & 5.65 & 92.6 & 3585 & 0.719 & 3.13 & 92.5 & 5529 & 0.710 & 4.99 \\
    \rowcolor{LightOrange}FoReaL-Decoding & $n{=}25,\,p{=}1.0$ & 66.7 & 11116 & 0.683 & 10.54 & 57.6 & 6065 & 0.882 & 6.68 & 94.5 & 3403 & 0.890 & 3.50 & 95.0 & 5422 & 0.872 & 5.66 \\
    \midrule
    \multicolumn{18}{l}{\textbf{DeepSeek-R1-Distill-Qwen-32B + Qwen2.5-1.5B-Instruct}}\\
    \midrule
    \multicolumn{2}{l}{DeepSeek-R1-Distill-Qwen-32B} & 66.7 & 13035 & - & 15.72 & 59.6 & 6602 & - & 8.09 & 93.6 & 3542 & - & 4.13 & 95.0 & 6243 & - & 7.54 \\
    \multicolumn{2}{l}{Qwen2.5-1.5B-Instruct} & 0.0 & 998 & - & 0.12 & 23.7 & 923 & - & 0.12 & 49.2 & 747 & - & 0.09 & 15.0 & 818 & - & 0.10 \\
    \midrule
    \rowcolor{LightBlue}FoReaL-Decoding & $n{=}15,\,p{=}0.8$ & 20.0 & 12584 & 0.571 & 9.05 & 47.5 & 7013 & 0.587 & 5.63 & 76.2 & 3792 & 0.614 & 2.85 & 65.0 & 7629 & 0.514 & 5.22 \\
    \rowcolor{LightBlue}FoReaL-Decoding & $n{=}15,\,p{=}1.0$ & 20.0 & 14188 & 0.588 & 11.19 & 47.5 & 6294 & 0.737 & 5.86 & 85.9 & 3894 & 0.750 & 3.28 & 65.0 & 7673 & 0.707 & 6.15 \\
    \rowcolor{LightOrange}FoReaL-Decoding & $n{=}25,\,p{=}0.8$ & 36.7 & 11575 & 0.710 & 9.58 & 56.7 & 4718 & 0.719 & 4.37 & 82.0 & 3025 & 0.729 & 2.52 & 72.5 & 5415 & 0.649 & 4.65 \\
    \rowcolor{LightOrange}FoReaL-Decoding & $n{=}25,\,p{=}1.0$ & 40.0 & 11239 & 0.813 & 11.00 & 57.1 & 5944 & 0.887 & 6.27 & 90.8 & 3403 & 0.894 & 3.36 & 92.5 & 6989 & 0.867 & 6.88 \\
    \midrule
    \multicolumn{18}{l}{\textbf{DeepSeek-R1-Distill-Qwen-1.5B + Qwen2.5-7B-Instruct}}\\
    \midrule
    \multicolumn{2}{l}{DeepSeek-R1-Distill-Qwen-1.5B} & 23.3 & 18021 & - & 2.86 & 22.2 & 8696 & - & 1.13 & 81.4 & 6704 & - & 1.14 & 65.0 & 13311 & - & 2.51 \\
    \multicolumn{2}{l}{Qwen2.5-7B-Instruct} & 6.7 & 1243 & - & 0.95 & 38.4 & 1054 & - & 0.89 & 76.0 & 773 & - & 0.61 & 52.5 & 994 & - & 0.75 \\
    \midrule
    \rowcolor{LightBlue}FoReaL-Decoding & $n{=}15,\,p{=}0.8$ & 16.7 & 4120 & 0.545 & 2.05 & 34.3 & 2130 & 0.602 & 1.07 & 76.4 & 1341 & 0.634 & 0.57 & 57.5 & 2515 & 0.580 & 1.08 \\
    \rowcolor{LightBlue}FoReaL-Decoding & $n{=}15,\,p{=}1.0$ & 16.7 & 14132 & 0.651 & 6.47 & 29.8 & 7913 & 0.703 & 3.08 & 79.6 & 3480 & 0.735 & 1.42 & 52.5 & 7330 & 0.686 & 3.35 \\
    \rowcolor{LightOrange}FoReaL-Decoding & $n{=}25,\,p{=}0.8$ & 20.0 & 4474 & 0.693 & 1.57 & 33.1 & 1801 & 0.718 & 0.80 & 78.6 & 1498 & 0.736 & 0.55 & 65.0 & 3778 & 0.683 & 1.76 \\
    \rowcolor{LightOrange}FoReaL-Decoding & $n{=}25,\,p{=}1.0$ & 23.3 & 11436 & 0.841 & 3.18 & 29.3 & 6800 & 0.863 & 2.53 & 79.2 & 3586 & 0.891 & 1.04 & 60.0 & 5721 & 0.865 & 1.66 \\
    \bottomrule
  \end{tabular}}%
  \label{tab:appendix_detailed_main}
\end{table}

\begin{table}[h]
  \caption{The detailed results of different collaborative settings on AIME24 and AMC23, including length and ratio.}
  \centering
  \resizebox{\textwidth}{!}{%
  \begin{tabular}{lc c c c c c c c c}
    \toprule
    \multicolumn{2}{c}{\textbf{Model}} &
    \multicolumn{4}{c}{\textbf{AIME24}} &
    \multicolumn{4}{c}{\textbf{AMC23}}\\
    \cmidrule(lr){1-2}\cmidrule(lr){3-6}\cmidrule(lr){7-10}
    \textbf{Method} & \textbf{Config} & \textbf{ACC (\%)} & \textbf{Length} & \textbf{Ratio} & \textbf{TFLOPs} & \textbf{ACC (\%)} & \textbf{Length} & \textbf{Ratio} & \textbf{TFLOPs} \\
    \midrule
    \multicolumn{10}{l}{\textbf{DeepSeek-R1-Distill-Qwen-32B + DeepSeek-R1-Distill-Qwen-1.5B}}\\
    \midrule
    \multicolumn{2}{l}{DeepSeek-R1-Distill-Qwen-32B} & 66.7 & 13035 & - & 15.72 & 95.0 & 6243 & - & 7.54 \\
    \multicolumn{2}{l}{DeepSeek-R1-Distill-Qwen-1.5B} & 23.3 & 18021 & - & 2.86 & 65.0 & 13311 & - & 2.51 \\
    \midrule
    \rowcolor{LightOrange}
    FoReaL-Decoding & $n{=}5,\,p{=}0.2$ & 23.3 & 12926 & 0.076 & 3.47 & 77.5 & 5634 & 0.089 & 1.39 \\
    \rowcolor{LightOrange}
    FoReaL-Decoding & $n{=}5,\,p{=}0.4$ & 26.7 & 11590 & 0.145 & 3.92 & 80.0 & 6549 & 0.157 & 2.18 \\
    \rowcolor{LightOrange}
    FoReaL-Decoding & $n{=}5,\,p{=}0.6$ & 36.7 & 11560 & 0.202 & 4.81 & 80.0 & 7081 & 0.228 & 2.95 \\
    \rowcolor{LightOrange}
    FoReaL-Decoding & $n{=}5,\,p{=}0.8$ & 43.3 & 11907 & 0.270 & 5.82 & 82.5 & 6399 & 0.294 & 3.29 \\
    \rowcolor{LightOrange}
    FoReaL-Decoding & $n{=}5,\,p{=}1.0$ & 50.0 & 13750 & 0.328 & 7.82 & 85.0 & 6916 & 0.355 & 3.86 \\
    \rowcolor{LightBlue}
    FoReaL-Decoding & $n{=}15,\,p{=}0.2$ & 26.7 & 12457 & 0.138 & 4.03 & 70.0 & 6680 & 0.154 & 2.05 \\
    \rowcolor{LightBlue}
    FoReaL-Decoding & $n{=}15,\,p{=}0.4$ & 33.3 & 11876 & 0.272 & 5.60 & 80.0 & 6043 & 0.303 & 2.91 \\
    \rowcolor{LightBlue}
    FoReaL-Decoding & $n{=}15,\,p{=}0.6$ & 50.0 & 10934 & 0.401 & 6.77 & 80.0 & 6460 & 0.429 & 3.99 \\
    \rowcolor{LightBlue}
    FoReaL-Decoding & $n{=}15,\,p{=}0.8$ & 50.0 & 11532 & 0.527 & 8.47 & 90.0 & 7037 & 0.571 & 5.37 \\
    \rowcolor{LightBlue}
    FoReaL-Decoding & $n{=}15,\,p{=}1.0$ & 66.7 & 10617 & 0.666 & 9.16 & 92.5 & 5942 & 0.708 & 5.28 \\
    \rowcolor{LightOrange}
    FoReaL-Decoding & $n{=}25,\,p{=}0.2$ & 36.7 & 10805 & 0.178 & 3.88 & 77.5 & 6798 & 0.193 & 2.32 \\
    \rowcolor{LightOrange}
    FoReaL-Decoding & $n{=}25,\,p{=}0.4$ & 33.3 & 11428 & 0.347 & 6.30 & 80.0 & 5929 & 0.362 & 3.17 \\
    \rowcolor{LightOrange}
    FoReaL-Decoding & $n{=}25,\,p{=}0.6$ & 50.0 & 10816 & 0.515 & 7.71 & 90.0 & 6169 & 0.537 & 4.49 \\
    \rowcolor{LightOrange}
    FoReaL-Decoding & $n{=}25,\,p{=}0.8$ & 53.3 & 12081 & 0.675 & 10.95 & 92.5 & 5529 & 0.710 & 4.99 \\
    \rowcolor{LightOrange}
    FoReaL-Decoding & $n{=}25,\,p{=}1.0$ & 66.7 & 11117 & 0.683 & 10.54 & 95.0 & 5422 & 0.872 & 5.66 \\
    \rowcolor{LightBlue}
    FoReaL-Decoding & $n{=}\infty,\,p{=}0.2$ & 30.0 & 12241 & 0.204 & 4.84 & 75.0 & 6502 & 0.216 & 2.43 \\
    \rowcolor{LightBlue}
    FoReaL-Decoding & $n{=}\infty,\,p{=}0.4$ & 46.7 & 11906 & 0.417 & 7.37 & 85.0 & 6719 & 0.423 & 4.07 \\
    \rowcolor{LightBlue}
    FoReaL-Decoding & $n{=}\infty,\,p{=}0.6$ & 50.0 & 11515 & 0.605 & 9.69 & 92.5 & 5671 & 0.607 & 4.42 \\
    \rowcolor{LightBlue}
    FoReaL-Decoding & $n{=}\infty,\,p{=}0.8$ & 60.0 & 10538 & 0.798 & 10.83 & 92.5 & 5925 & 0.797 & 5.87 \\
    \rowcolor{LightBlue}
    FoReaL-Decoding & $n{=}\infty,\,p{=}1.0$ & 66.7 & 13035 & 1.000 & 15.72 & 95.0 & 6244 & 1.000 & 7.54 \\
    
    \midrule
    \multicolumn{10}{l}{\textbf{DeepSeek-R1-Distill-Qwen-1.5B + Qwen2.5-7B-Instruct}}\\
    \midrule
    \multicolumn{2}{l}{DeepSeek-R1-Distill-Qwen-1.5B} & 23.3 & 18021 & - & 2.86 & 65.0 & 13311 & - & 2.51 \\
    \multicolumn{2}{l}{Qwen2.5-7B-Instruct} & 6.7 & 1243 & - & 0.95 & 52.5 & 994 & - & 0.75 \\
    \midrule
    \rowcolor{LightOrange}
    FoReaL-Decoding & $n{=}5,\,p{=}0.2$ & 10.0 & 1047 & 0.170 & 0.73 & 50.0 & 923 & 0.179 & 0.64 \\
    \rowcolor{LightOrange}
    FoReaL-Decoding & $n{=}5,\,p{=}0.4$ & 10.0 & 1381 & 0.230 & 0.91 & 55.0 & 1065 & 0.244 & 0.69 \\
    \rowcolor{LightOrange}
    FoReaL-Decoding & $n{=}5,\,p{=}0.6$ & 13.3 & 2377 & 0.306 & 1.61 & 62.5 & 2574 & 0.302 & 1.97 \\
    \rowcolor{LightOrange}
    FoReaL-Decoding & $n{=}5,\,p{=}0.8$ & 13.3 & 4203 & 0.345 & 2.87 & 47.5 & 2897 & 0.373 & 1.83 \\
    \rowcolor{LightOrange}
    FoReaL-Decoding & $n{=}5,\,p{=}1.0$ & 16.7 & 7236 & 0.382 & 4.45 & 50.0 & 5614 & 0.428 & 3.24 \\
    \rowcolor{LightBlue}
    FoReaL-Decoding & $n{=}15,\,p{=}0.2$ & 3.3 & 1936 & 0.208 & 1.38 & 47.5 & 985 & 0.224 & 0.65 \\
    \rowcolor{LightBlue}
    FoReaL-Decoding & $n{=}15,\,p{=}0.4$ & 16.7 & 1189 & 0.339 & 0.70 & 45.0 & 1055 & 0.360 & 0.61 \\
    \rowcolor{LightBlue}
    FoReaL-Decoding & $n{=}15,\,p{=}0.6$ & 16.7 & 1793 & 0.455 & 0.92 & 52.5 & 1307 & 0.482 & 0.65 \\
    \rowcolor{LightBlue}
    FoReaL-Decoding & $n{=}15,\,p{=}0.8$ & 16.7 & 4120 & 0.545 & 2.05 & 57.5 & 2515 & 0.580 & 1.08 \\
    \rowcolor{LightBlue}
    FoReaL-Decoding & $n{=}15,\,p{=}1.0$ & 16.7 & 14132 & 0.651 & 6.47 & 52.5 & 7330 & 0.686 & 3.35 \\
    \rowcolor{LightOrange}
    FoReaL-Decoding & $n{=}25,\,p{=}0.2$ & 13.3 & 1243 & 0.249 & 0.80 & 50.0 & 958 & 0.231 & 0.62 \\
    \rowcolor{LightOrange}
    FoReaL-Decoding & $n{=}25,\,p{=}0.4$ & 20.0 & 1317 & 0.389 & 0.73 & 42.5 & 1077 & 0.405 & 0.59 \\
    \rowcolor{LightOrange}
    FoReaL-Decoding & $n{=}25,\,p{=}0.6$ & 16.7 & 1743 & 0.536 & 0.79 & 57.5 & 2047 & 0.560 & 1.14 \\
    \rowcolor{LightOrange}
    FoReaL-Decoding & $n{=}25,\,p{=}0.8$ & 20.0 & 4474 & 0.693 & 1.57 & 65.0 & 3778 & 0.683 & 1.76 \\
    \rowcolor{LightOrange}
    FoReaL-Decoding & $n{=}25,\,p{=}1.0$ & 23.3 & 11436 & 0.841 & 3.18 & 65.0 & 5721 & 0.865 & 1.66 \\
    \rowcolor{LightBlue}
    FoReaL-Decoding & $n{=}\infty,\,p{=}0.2$ & 13.3 & 1072 & 0.260 & 0.69 & 50.0 & 986 & 0.290 & 0.61 \\
    \rowcolor{LightBlue}
    FoReaL-Decoding & $n{=}\infty,\,p{=}0.4$ & 6.7 & 1276 & 0.420 & 0.68 & 42.5 & 1140 & 0.467 & 0.56 \\
    \rowcolor{LightBlue}
    FoReaL-Decoding & $n{=}\infty,\,p{=}0.6$ & 10.0 & 1914 & 0.614 & 0.78 & 57.5 & 1324 & 0.618 & 0.53 \\
    \rowcolor{LightBlue}
    FoReaL-Decoding & $n{=}\infty,\,p{=}0.8$ & 23.3 & 4244 & 0.788 & 1.26 & 65.0 & 2854 & 0.817 & 0.79 \\
    \rowcolor{LightBlue}
    FoReaL-Decoding & $n{=}\infty,\,p{=}1.0$ & 23.3 & 18021 & 1.000 & 2.86 & 65.0 & 13311 & 1.000 & 2.51 \\
    \bottomrule
  \end{tabular}}%
  \label{tab:appendix_detailed_tradeoff}
\end{table}



\end{document}